# Title: Latent Space Data Fusion Outperforms Early Fusion in Multimodal Mental Health Digital Phenotyping Data

Authors: Youcef Barkat, Dylan Hamitouche, Deven Parekh, Ivy Guo, David Benrimoh

Douglas Research Centre, McGill University, Verdun, Canada

## Abstract

**Background**

Mental illnesses such as depression and anxiety require improved methods for early detection and personalized intervention. Traditional predictive models often rely on unimodal data or early fusion strategies that fail to capture the complex, multimodal nature of psychiatric data. Advanced integration techniques, such as intermediate (latent space) fusion, may offer better accuracy and clinical utility.

**Methods**

Using data from the BRIGHTEN clinical trial, we evaluated intermediate (latent space) fusion for predicting daily depressive symptoms (PHQ-2 scores). We compared early fusion implemented with a Random Forest (RF) model and intermediate fusion implemented via a Combined Model (CM) using autoencoders and a neural network. The dataset included behavioral (smartphone-based), demographic, and clinical features. Experiments were conducted across multiple temporal splits and data stream combinations. Performance was evaluated using mean squared error (MSE) and coefficient of determination ($R^2$).

**Results**

The CM outperformed both RF and Linear Regression (LR) baselines across all setups, achieving lower MSE (0.4985 vs. 0.5305 with RF) and higher $R^2$ (0.4695 vs. 0.4356). The RF model showed signs of overfitting, with a large gap between training and test performance, while the CM maintained consistent generalization. Performance was best when integrating all data modalities in the CM (in contradistinction to RF), underscoring the value of latent space fusion for capturing non-linear interactions in complex psychiatric datasets.

**Conclusion**

Latent space fusion offers a robust alternative to traditional fusion methods for prediction with multimodal mental health data. Future work should explore model interpretability and individual-level prediction for clinical deployment.

# Introduction

Mental health disorders, including depression and anxiety, present a significant public health challenge, necessitating improved methods for early detection and personalized intervention [1]. Traditional clinical assessments rely heavily on subjective self-reports and structured interviews, which are not only prone to biases and inconsistencies but are also time-consuming and costly to administer [2]. Consequently, there has been a growing interest in leveraging digital biomarkers—measurable indicators collected from digital devices, such as smartphones, wearables, and social media activity— derived from multimodal data sources to enhance predictive accuracy in mental health assessments. Multimodal data, which may include, among other data types, behavioral, digital, physiological, and psychometric assessments, offers a promising avenue for improving psychiatric outcome prediction [3,4]. However, integrating such diverse data types presents significant challenges, particularly in terms of scale, format, and temporal alignment [5]. Behavioral data, such as GPS location and accelerometer readings, are continuous and high frequency, while psychological assessments, such as PHQ-9 (Patient Health Questionnaire-9) scores are often categorical and collected relatively infrequently [6]. This scale disparity can lead to imbalances in model weighting and feature importance. Additionally, differences in format—such as structured physiological signals versus unstructured text from self-reports—complicate direct integration. Finally, temporal misalignment poses a major hurdle, as multimodal data streams are rarely synchronized, making it difficult to capture meaningful interactions between behavioral patterns and mental health states [5]. Indeed, prior work by Pratap et al. demonstrated that the best-performing model for predicting depression scores utilized only a subset of the available data modalities [7]. This study employed an early fusion approach, raising the question of whether an intermediate fusion strategy could enable more effective integration of all modalities to enhance predictive accuracy.

Fusion techniques are employed to integrate multimodal data while preserving meaningful interactions between different sources. Several fusion strategies have been developed, each with distinct advantages and limitations. Early fusion involves concatenating raw features from multiple modalities at the input level before passing them into a predictive model. While this method can capture statistical correlations between modalities, it often struggles with scale differences and redundant information. Late fusion, on the other hand, processes each modality independently and merges their outputs at the decision level, allowing for greater flexibility but potentially losing cross-modal dependencies [8]. Studies comparing early and late fusion strategies in multimodal learning suggest that the effectiveness of each approach depends on the data and task at hand. For instance, Gadzicki et al. (2022) evaluated the performance of early and late fusion in multimodal convolutional neural networks for human activity recognition. Their findings indicated that early fusion strategies could better exploit statistical correlations between different modalities, leading to superior performance in classification tasks compared to late fusion methods, which treat each modality independently before merging the final predictions [9]. Additionally, Krushnasamy & Rashinkar (2017) provided a broader review of data fusion techniques, emphasizing the benefits of integrating multiple data sources for improved prediction accuracy and robustness [10].

Several studies have explored smartphone-based passive data collection for predicting psychiatric outcomes [11]. Pratap et al. (2018) examined the use of passive phone sensor data, including GPS tracking, call and message logs, and PHQ-9 scores, to predict daily mood using a Random Forest model within the BRIGHTEN trial [11]. However, the study revealed key limitations of early fusion: while their unimodal model achieved good performance, the direct concatenation of features in early fusion failed to capture complex interactions between modalities. Indeed, their best performing model at the population level was one that did not include all possible data modalities available.

Beyond smartphone-based data, other studies have explored multimodal fusion techniques incorporating physiological signals. For instance, Li & Sano (2019) evaluated stress and mood prediction models by integrating physiological data from wearables with behavioral metrics, comparing early and late fusion techniques. Their findings indicated that while early fusion yielded better performance in some cases, it struggled with feature redundancy and noise, highlighting the need for more advanced fusion strategies [12].

More recently, intermediate fusion (also known as latent space fusion) has emerged as a promising approach, mapping each modality into a shared latent representation before predictive modeling [13]. This approach allows different data types to be projected into a common feature space where their underlying relationships can be captured more effectively. By encoding multimodal data into a structured latent space, the model can learn meaningful interactions between features while reducing noise and redundancy [14]. Furthermore, latent space fusion enhances generalization by enabling the model to focus on abstract representations of the data rather than being constrained by raw feature disparities.

Building on prior research, this study aims to validate latent space fusion techniques using the BRIGHTEN dataset, a multimodal dataset collected through remote clinical trials [11]. By comparing early and latent space fusion within a group-level modeling framework, we assess their impact on mental health data prediction and examine the contribution of individual data modalities, such as behavioral patterns and PHQ-9 scores, to model performance. Through this work, we extend prior findings by integrating advanced fusion strategies and reinforcing the important role of multimodal data integration in computational psychiatry. By consistently outperforming traditional fusion methods in predictive accuracy and generalization, the latent state approach offers a promising pathway toward improving early detection of clinically significant symptom changes, laying the groundwork for more timely and personalized mental health interventions.

# Methods

## 1. Data Description

The BRIGHTEN dataset originates from two fully remote smartphone-based clinical trials—BRIGHTEN V1 and BRIGHTEN V2—which evaluated digital interventions for depression. Conducted across the United States, these trials enrolled 2,193 participants, generating a diverse and extensive multimodal dataset for psychiatric research [11].

This dataset consists of three primary data types: self-reported psychological assessments, passive phone-based behavioral data, and demographic information. The psychological assessments include daily mood ratings based on the Patient Health Questionnaire-2 (PHQ-2) and weekly PHQ-9 scores, which track depressive symptoms over time [11]. Passive data streams, collected via smartphone sensors, capture mobility patterns (e.g., distance traveled, movement radius, and location variability based on GPS), phone usage metrics (e.g., call duration, SMS count, and missed interactions), and social interaction diversity (e.g., the number of unique contacts engaged through calls and messages) [11]. These passive indicators provide objective measures of social engagement, physical activity, and behavioral patterns, offering insights into mood fluctuations and depression severity. Given the variability in individual phone usage and movement behaviors, this dataset enables personalized mood prediction models, helping to assess the feasibility of passive data for mental health monitoring [11].

In this study, we utilize the original BRIGHTEN V1 dataset, as used by Pratap et al. (2018). To ensure a direct comparison with prior research, we apply the same inclusion criteria as Pratap et al. (2018), focusing on participants with sufficient data availability across all relevant modalities [11].

## 2. Data Preprocessing

To ensure the reliability and consistency of the multimodal dataset, a structured preprocessing pipeline was implemented. Given the heterogeneity in data types within the BRIGHTEN dataset, several key steps were undertaken to clean, standardize, and integrate the data before model training.

A major challenge in working with real-world multimodal data is the presence of missing values, which can arise due to user compliance issues, sensor failures, or incomplete survey responses. Missing values were first identified across different subsets of the dataset, including passive phone-based features (e.g., call logs, GPS activity, interaction diversity), demographic data (e.g., age, gender, marital status), baseline PHQ-9 scores, and daily PHQ-2 assessments, which serve as the primary outcome variable. To address these gaps, we employed MissForest, a robust iterative imputation method based on Random Forests, which has been shown to perform well for datasets containing both categorical and continuous variables [15].

Once missing values were handled, the different data modalities were merged into a unified feature set. Passive features were first combined with demographic data using participant ID as a unique identifier, followed by the addition of baseline PHQ-9 scores. The daily PHQ-2 scores were then concatenated to ensure temporal alignment between the predictor variables and the outcome measure. Given that different modalities may contribute uniquely to predictive modeling, multiple feature subsets were created for further analysis. These include a passive-only dataset focusing on smartphone-based behavior, a combination of demographic and passive features, a set integrating demographic and PHQ-9 clinical self-reports, and a full dataset incorporating all available features.

To account for variations in scale across different data types, feature standardization was applied using the StandardScaler from scikit-learn [16]. Standardization ensures that numerical features are centered around zero with unit variance, preventing variables with larger magnitudes from disproportionately influencing the learning process [17]. Additionally, categorical demographic variables such as gender and marital status were converted into numerical representations through one-hot encoding to ensure compatibility with machine learning models.

By implementing this preprocessing pipeline, we optimized the multimodal dataset for predictive modeling, reducing the impact of missing values while preserving key behavioral and psychological patterns. This approach ensures that the data used for modeling is structured in a way that maximizes predictive performance and generalization potential.

## 3. Fusion Techniques

The integration of multimodal data in computational psychiatry poses challenges due to differences in scale, format, and temporal resolution across data sources [18]. To address these complexities, this study evaluates two fusion approaches: early fusion, which directly concatenates raw features, and latent space fusion, which maps heterogeneous data into a shared representation before predictive modeling.

While early fusion is simple and computationally efficient, it assumes equal contribution from all modalities and struggles with redundancy, noise, and non-linear interactions. Additionally, concatenating structured demographic variables with high-dimensional time-series behavioral data can lead to overfitting and poor generalization, as observed in cases where early fusion failed to improve performance beyond unimodal models[19][20].

To overcome these limitations, latent space fusion (also referred to as intermediate fusion) has emerged as a more robust alternative. In this study, we employ an autoencoder-based latent space fusion framework [21], where an encoding phase compresses raw data into lower-dimensional representations, preserving key information while filtering out noise. The target variable for prediction in our models, as in the Pratap study, is the PHQ-2 score, which provides a brief measure of depressive symptom severity. These latent variables are then processed by a neural regression model to predict PHQ-2 scores.

One of the key advantages of latent space fusion is its ability to capture non-linear interactions between different data modalities. Unlike early fusion, latent space fusion allows the model to learn meaningful relationships between behavioral, demographic, and clinical variables. This is particularly important in psychiatry, where mental health states are influenced by a combination of biological, environmental, and behavioral factors that may not have simple additive effects. Additionally, by compressing raw features into latent variables, this method reduces dimensionality, mitigating issues related to high-dimensional data and ensuring that only the most informative aspects of each modality are retained. Furthermore, latent space fusion improves the generalization ability of predictive models by reducing overfitting to modality-specific noise [22].

## 4. Modeling Approaches

To assess the effectiveness of different fusion strategies for daily mood prediction, this study employs two primary modeling approaches: Random Forest (RF) with early fusion and a neural network-based Combined Model (CM) with latent space fusion. Each approach is designed to evaluate how different data integration strategies influence predictive performance, particularly in handling complex multimodal psychiatric data.

The Random Forest (RF) model serves as a baseline approach and follows the methodology established in previous studies, such as Pratap et al. (2018) [11]. In this approach, multimodal data including behavioral, demographic, and psychological assessments are concatenated at the feature level before being used as inputs for the model. The RF model is an ensemble learning technique that constructs multiple decision trees and aggregates their predictions to improve accuracy and reduce variance [23]; the prediction target was the PHQ-2 score.

The CM integrates multimodal data using latent space fusion. The CM consists of two key components: an autoencoder-based feature extractor and a neural network regressor (Fig. 1). The autoencoder compresses input data from each modality into a shared latent space, capturing essential patterns while filtering out irrelevant noise. These latent representations are then fed into a fully connected neural network, which predicts PHQ-2 scores as an indicator of daily mood fluctuations.

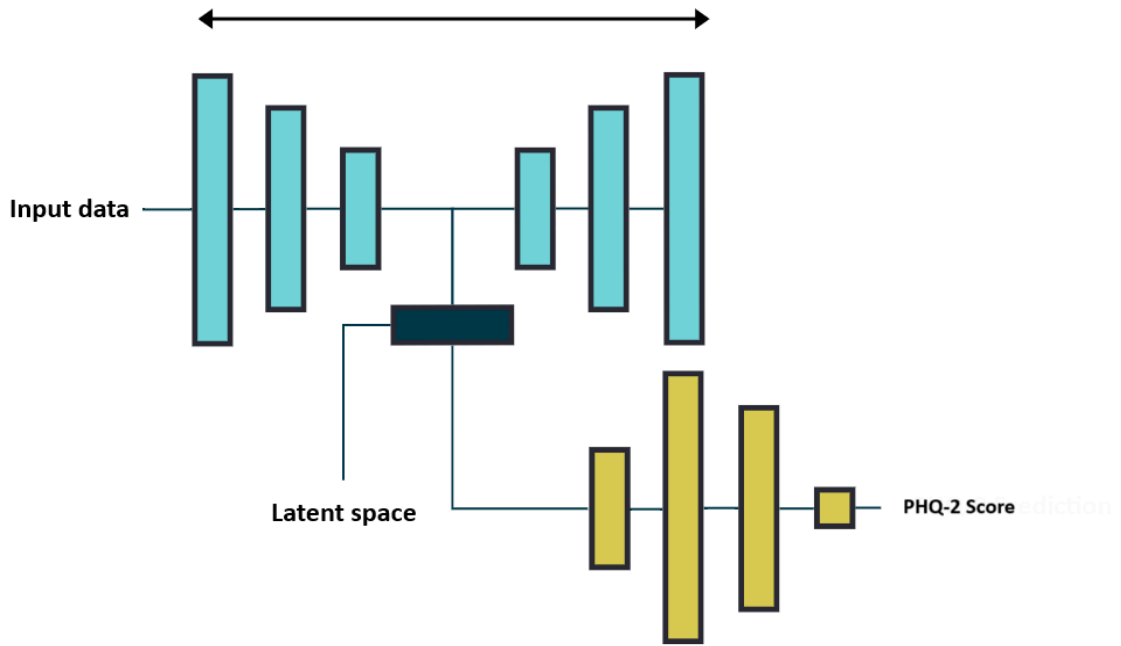

*Figure 1 : Detailed Structure of the Combined Model with Data Propagation from the Autoencoder to the Neural Network for PHQ-2 Score Prediction.*

Additionally, the latent space representation enhances model robustness by reducing dimensionality and improving feature integration across modalities. Since each modality is encoded separately before fusion, the model can better capture complementary information from different sources. This flexibility is crucial for real-world psychiatric data, which often come from heterogeneous modalities collected at different frequencies.

Overall, the modeling approaches investigated in this study reflect a shift toward more advanced multimodal fusion techniques in computational psychiatry. While Random Forest (RF) remains a widely used machine learning approach, it is inherently incompatible with latent space fusion, as it cannot process learned latent representations directly. In contrast, the Combined Model (CM) with latent space fusion provides a more flexible and adaptive learning framework, positioning it as an alternative for future psychiatric outcome prediction models when sufficient multimodal data are available.

## 5. Experimental Setup

The dataset was partitioned into training and test sets following the same approach as Pratap et al. (2018) [11]. The target variable in our predictive models was PHQ-2 scores, representing depressive symptom severity. The training set included all available data from the first four weeks of the study, while the remaining 8 weeks were allocated to the test set. This design allowed us to evaluate how well each model generalized to unseen PHQ-2 scores while maintaining temporal consistency. Additionally, we introduced a second evaluation strategy, where test data were sampled randomly from all weeks, ensuring robustness against varying temporal structures and potential distribution shifts. As an additional baseline, we also included a linear regression model for comparison to assess how more complex models performed relative to a simpler alternative. Each model was optimized using Mean Squared Error (MSE) as the loss function, with hyperparameters tuned using a grid search approach. Model performance was assessed using MSE and the Coefficient of Determination ($R^2$), two widely used metrics for evaluating regression models. To account for variability and ensure robustness of results, we repeated each experiment five times. By comparing these metrics across different experimental setups, we aimed to quantify the benefits of latent space fusion over early fusion techniques in terms of predictive accuracy and generalization ability.

# Results

For the entire study, we selected **131 participants** from the Brighten dataset based on data availability across all required modalities. On average, each participant contributed **36.03 days of data**, providing a substantial longitudinal window for mood prediction and model evaluation.

**Comparison of Early Fusion Random Forest and Latent Space Fusion Models**

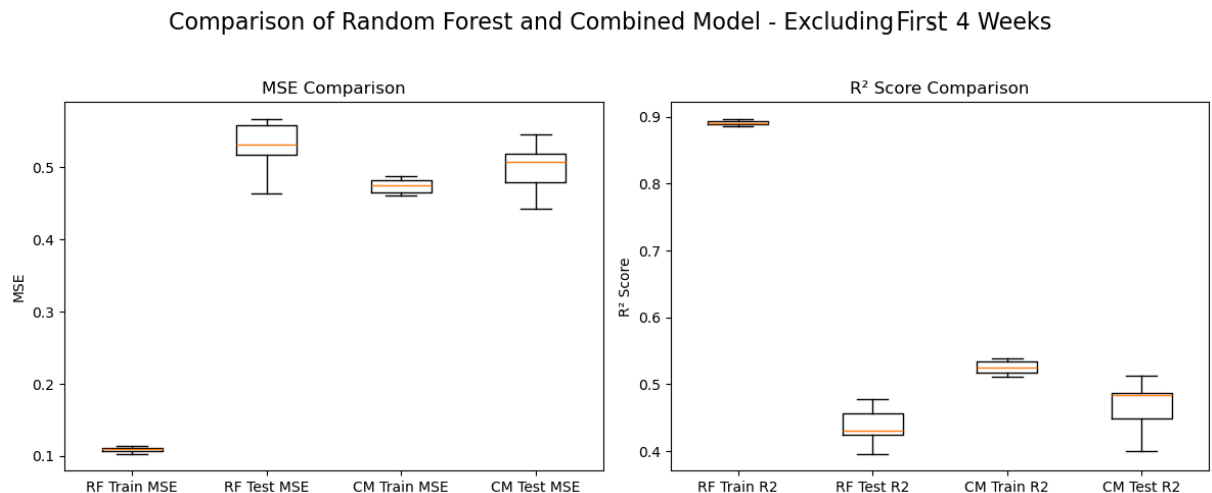

*Figure 2: Comparison of the Performance of Random Forest (RF) and Combined Model (CM) on MSE and R² Metrics, Excluding the first Four Weeks from the training set.*

Fig. 2 shows the comparison of the performance of the RF Model and CM on the test set, excluding the first four weeks of training data, using all available data modalities. Results demonstrated a clear advantage for the Combined Model (CM). . The CM achieved a test MSE of 0.4985, compared to 0.5305 for the Random Forest (RF) mode. In terms of $R^2$, the CM outperformed RF with a test score of 0.4695 versus 0.4356. Moreover, while RF achieved high training scores, the notable drop in test performance highlighted its tendency to overfit. In contrast, the CM maintained a more balanced performance across training and test sets, reinforcing its robustness and superior generalization.

**Incorporating All Weeks for Evaluation**

To further test the robustness of our approach, we conducted an additional experiment where test samples were drawn randomly from all weeks rather than following a strict chronological split, ensuring they were not included in the training data. This methodology allowed us to examine whether the CM maintained its predictive advantage across varied temporal structures, providing a more comprehensive assessment of its generalization capabilities (Fig. 3).

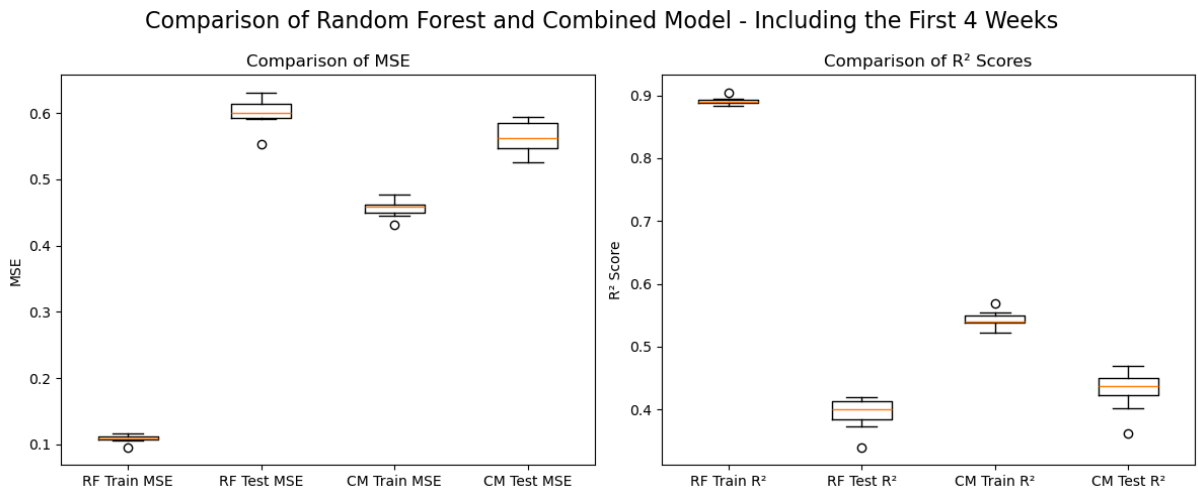

*Figure 3: Comparison of Random Forest (RF) and Combined Model (CM) on MSE and R² Metrics with Inclusion of All Weeks from the Dataset.*

The CM again exhibited a lower average test MSE of **0.5635** compared to **0.6007** for the RF, reaffirming its ability to generalize to unseen data. Similarly, the CM's test R² score of **0.4316** exceeded the RF's **0.3943**, further validating its improved predictive performance.

**Impact of Training Duration on Model Performance**

To investigate how the amount of training data impacts model effectiveness, we varied the number of weeks included in the training set and observed the evolution of MSE and R² scores for the CM.

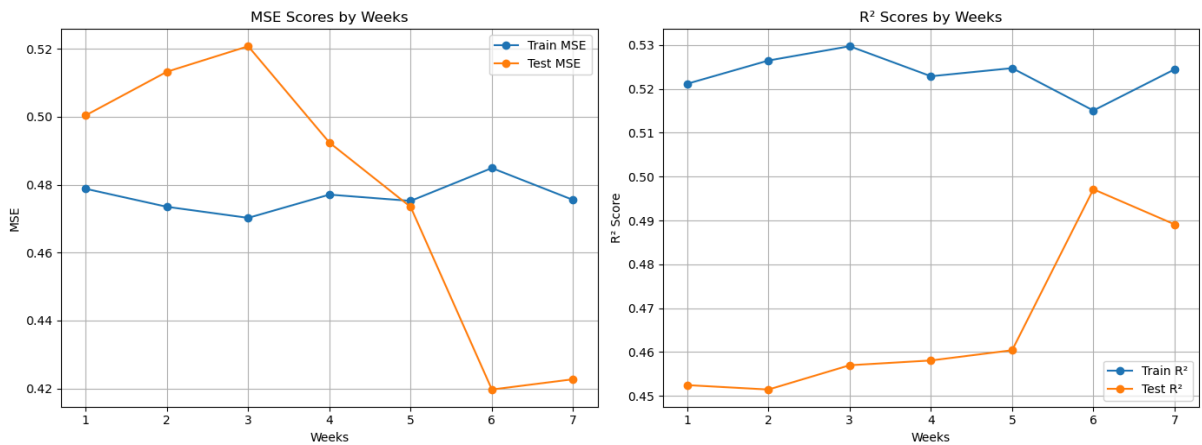

*Figure 4 : Weekly Variation of MSE and R² Metrics for the Combined Model Based on the Weeks Included in the Test Set.*

Results(Fig. 4) revealed that increasing the number of training weeks led to improved test performance**.** Specifically, while training MSE remained relatively stable between **0.46** and **0.48**, test MSE showed a substantial decline, dropping from **0.52** to approximately **0.44**, suggesting enhanced model generalization. Additionally, $R^2$ scores for training remained consistent above **0.50**, whereas test $R^2$ exhibited an increase, reaching up to **0.53**, highlighting the benefits of a longer training period.

**Comparison between combined model and linear regression**

To further assess the robustness of our modeling approach, we compared the CM with a Linear Regression (LR) model employing early fusion, hypothesizing that LR would be less prone to overfitting than RF. Unlike RF, which exhibited signs of overfitting, Linear Regression serves as a simpler baseline that prioritizes interpretability and generalizability. This comparison allowed us to evaluate whether the CM provided a meaningful improvement over a more conventional, well-regularized model.

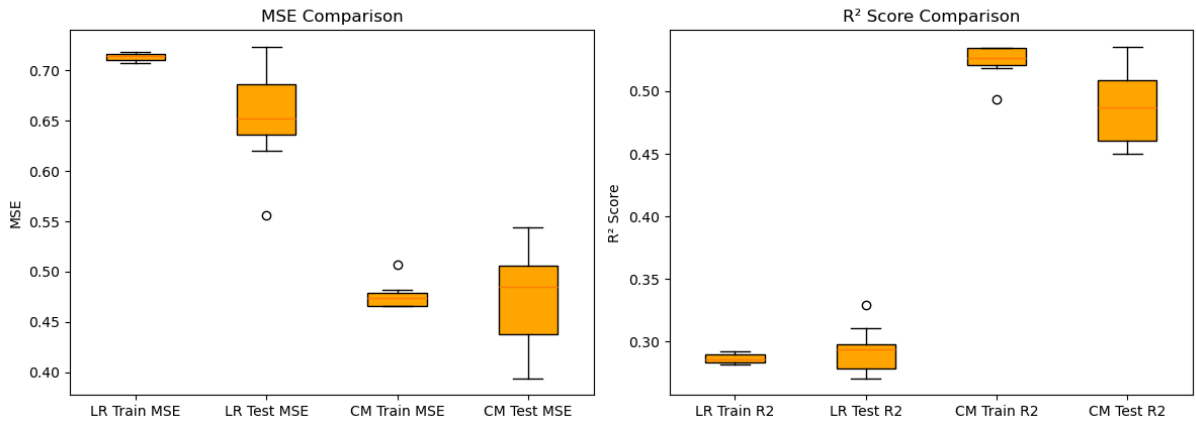

*Figure 5: Comparison of the Combined Model (CM) and Linear Regression Model (LR) on MSE and R² Metrics, Excluding the First Four Weeks of the Test Set.*

Results (Fig. 5) indicate that **CM consistently outperforms LR**. Specifically, in terms of MSE, the LR model achieved a **test MSE of 0.65**, whereas the CM demonstrated **lower error rates with a test MSE of 0.4985**. Similarly, the R² score for LR was **0.29 on the test set**, indicating limited ability to explain variance in the data. In contrast, the CM achieved a **test R² of 0.4695**, showcasing a stronger ability to capture the complexity of multimodal features.

**Performance Across Different Modalities and Modality Combinations**

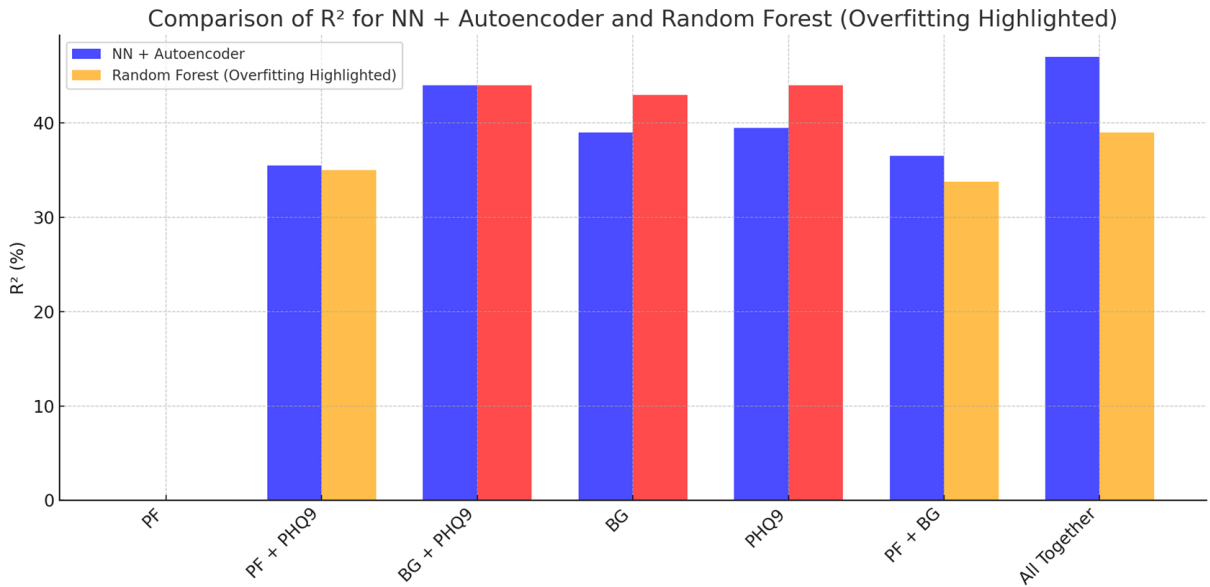

*Figure 6: Comparison of R² Scores for Combined Model (NN + Autoencoder) and Random Forest (RF) Across Different Feature Modality Combinations, RF Highlighting Overfitting (RED).*

To investigate the contribution of different modalities to PHQ-2 prediction performance, we evaluated our CM and RF models using various combinations of passive features (PF), background demographics (BG), and past PHQ-9 scores. Results, presented in figure 6, illustrate the impact of each modality on R² scores. Across all tested configurations, the CM consistently outperformed RF, particularly in settings that combined multiple modalities. The highest performance gain was observed when integrating all available data sources — a notable contrast to Pratap et al. (2018), where strongest performance was achieved using unimodal data alone, reinforcing the importance of multimodal fusion for optimizing the use of rich multimodal datasets. Additionally, RF's tendency to overfit is evident, whereas the CM maintained consistent generalization, demonstrating its robustness in handling complex multimodal interactions.

# Discussion

The findings of this study highlight the advantages of latent space fusion in improving the predictive accuracy of psychiatric outcome models. Compared to traditional early fusion techniques, our approach consistently demonstrated lower MSE and higher R² scores across various experimental setups. The robustness of the CM was further validated through different data splits, showcasing its ability to generalize effectively to unseen data. Most importantly, the model achieved its highest performance when all data modalities were integrated (in contradistinction to RF), indicating that latent space fusion can more fully harness the potential of multimodal inputs to deliver more accurate and reliable mental health predictions.

A key takeaway from this study is the clear limitation of traditional data fusion approaches, which struggle with the direct concatenation of heterogeneous data types. This limitation is exacerbated when dealing with complex multimodal data, such as behavioral metrics, self-reported assessments, and social activity logs. In contrast, the latent space approach mitigates these challenges by transforming raw features into a shared representation before predictive modeling, reducing noise and enhancing feature extraction. The results reinforce the idea that an intermediate fusion strategy is more suitable for capturing intricate relationships between different data modalities.

Furthermore, our comparison between the CM and LR was essential for evaluating model generalization. While RF showed signs of overfitting, Linear Regression provided a simpler, less prone to overfitting, and more interpretable baseline. Despite this the CM still consistently outperformed LR, highlighting its ability to capture nonlinear dependencies in the data. Although neural networks are often criticized for their lack of interpretability, certain techniques—such as feature importance scores, saliency mapping, or layer-wise relevance propagation—can help partially mitigate this issue. Nonetheless, interpretability remains a known limitation of neural network models, particularly in sensitive applications like psychiatry. Prior work has explored ways to balance predictive power and interpretability in multimodal contexts, suggesting a continued need for methodological development in this area [25] [26] [27] [28].

Future work should focus on developing explainability tools to help clinicians understand how different data modalities contribute to model predictions[25]. Additionally, integrating physiological signals, such as heart rate variability and electrodermal activity, may further enhance model performance by capturing physiological correlates of psychiatric states [29].

This study has several limitations. One limitation relates to data availability and collection. Although the BRIGHTEN dataset provides a rich source of multimodal data, real-world implementations would require continuous and scalable data collection frameworks. Addressing missing data and ensuring reliable passive sensing in diverse populations will be critical for translating these models into clinical practice. In addition, this study focused on group-level modeling; previous work has shown that individual-level modelling may be an effective approach when analysing digital phenotyping data [11], and we have not in this work applied latent fusion to individual level models., Future planned research will investigate how these techniques can be adapted for personalized predictions. Tailoring models to individual patients by incorporating personal behavioral baselines and contextual factors may lead to more precise and actionable insights. By bridging computational psychiatry with real-world applications, advanced fusion techniques have the potential to significantly improve mental health predictions and therefore assessment and intervention strategies.

**Funding:** This research was made possible through the generous support of the Emerging Challenges Modelling Project, an initiative led by the Centre de Recherches Mathématiques (CRM) in partnership with GERAD and UNIQUE, and funded by the Quebec Research Fund (FRQ); the Strategia Program at the CRM; the McGill Computational and Data Systems Initiative. Additional funding and support were provided by the Douglas Research Center Startup Funds, the Fonds de recherche du Québec Junior 1 award (DB), and the Healthy Brains, Healthy Lives (HBHL) initiative at McGill University. Dr. Benrimoh has also received funding from a Brain & Behavior Research Foundation Young Investigator Award.

Author contributions: YB and DB conceptualized the study. YB ran the analyses, and wrote the first draft of the paper. DB, DH, DP, and IG all reviewed and edited the paper. DB provided supervision.